\documentclass[letterpaper, 10pt, conference]{ieeeconf}

\IEEEoverridecommandlockouts
\makeatletter
\let\NAT@parse\undefined
\makeatother
\usepackage[numbers, sort]{natbib}

\usepackage{amsmath}
\usepackage{amssymb}
\usepackage{booktabs}
\usepackage{siunitx}
\usepackage{graphicx}
\usepackage{subfigure}
\usepackage{algorithm}
\usepackage{algorithmic}[1]
\usepackage{color}
\usepackage{array}
\usepackage{url}
\usepackage{flushend}
\usepackage{lipsum}
\usepackage{makecell}
\usepackage{multirow}
\usepackage{hyperref}
\def\secref#1{Section~\ref{#1}}
\def\figref#1{Figure~\ref{#1}}
\def\tabref#1{Table~\ref{#1}}
\def\eqref#1{Eq.~\ref{#1}}

\newcommand{\bZero}{\mathbf{0}}

\newcommand{\bd}{\mathbf{d}}

\newcommand{\bzh}{\mathbf{\hat{z}}}

\newcommand{\bF}{\mathbf{F}}
\newcommand{\bG}{\mathbf{G}}

\newcommand{\bp}{\mathbf{p}}
\newcommand{\by}{\mathbf{y}}

\newcommand{\be}{\mathbf{e}}

\newcommand{\bx}{\mathbf{x}}

\newcommand{\bz}{\mathbf{z}}

\newcommand{\bSigma}{\mathbf{\Sigma}}

\def\argmin{\mathop{\rm argmin}}



\newcommand{\degree}{^{\circ}}
\newcommand{\seqset}{\mathcal{S}}
\newcommand{\seqsize}{S}

\def\argmin{\mathop{\rm argmin}}

\title{\huge Metric Localization using Google Street View}

\author{Pratik Agarwal \and Wolfram Burgard \and Luciano Spinello
  \thanks{\scriptsize All authors are with the University of Freiburg,
    Institue of Computer Science, 79110 Freiburg, Germany.
	This work has been partially supported by BMBF under contract number
13EZ1129B-iView and by the EC under contract number ERG-AG-PE7-267686-LifeNav and FP7-610603-EUROPA2.}}
\begin{document}
\maketitle

\begin{abstract}

  Accurate metrical localization is one of the central challenges in
  mobile robotics. Many existing methods aim at localizing after
  building a map with the robot.  In this paper, we present a novel
  approach that instead uses geotagged panoramas from the Google
  Street View as a source of global positioning.  We model the problem
  of localization as a non-linear least squares estimation in two
  phases.  The first estimates the 3D position of tracked feature
  points from short monocular camera sequences. The second computes
  the rigid body transformation between the Street View panoramas and
  the estimated points. The only input of this approach is a stream of
  monocular camera images and odometry estimates.  We quantified the
  accuracy of the method by running the approach on a robotic platform
  in a parking lot by using visual fiducials as ground truth.
  Additionally, we applied the approach in the context of personal
  localization in a real urban scenario by using data from a Google
  Tango tablet.

\end{abstract}

\section{Introduction}

Accurate metrical positioning is a key enabler for a set of crucial
applications, from autonomous robot navigation, intelligent driving
assistance to mobile robot localization systems.  During the past
years, the robotics and the computer vision community formulated
accurate localization solutions that model the localization problem as
pose estimation in a map generated with a robot. Given the importance
of map building, researchers have devoted significant resources on
building robust mapping methods~\cite{thrun2002a,dellaert2005rss,
  triggs2000bundle,Kaess12ijrr}.  Unfortunately, localization based on
maps built with robots still presents disadvantages. Firstly, it is
time consuming and expensive to compute an accurate map. Secondly, the
robot has to visit the environment beforehand. An alternative solution
is to re-use maps for localization even if they were not designed for
robots.  If maps do not exist, we can use various robot mapping
algorithms, but if maps exist and a robot can utilize them, it will
allow the robot to navigate without needing to explore the full environment
beforehand.

In this paper, we propose a novel approach that allows robots to localize
with maps built for humans for the purpose of visualizing places. Our
method does not require the construction of a new consistent map and nor does it
require the robot to previously visit the environment. Our central idea is to
leverage Google Street View as an abundant source of accurate geotagged imagery.
In particular, our key contribution is to formulate localization as the problem
of estimating the position of Street View's panoramic imagery relative to
monocular image sequences obtained from a moving camera.  With our approach, we
can leverage Google's global panoramic image database, with data collected each
5-10m and continuously updated across five continents~\cite{google-sv,
anguelov10google}.  To make this approach as general as possible, we only make
use of  a monocular camera and a metric odometry estimate, such as the one
computed from IMUs or wheel encoders.

\begin{figure}
\centering
\subfigure{\includegraphics[width=0.42\textwidth]{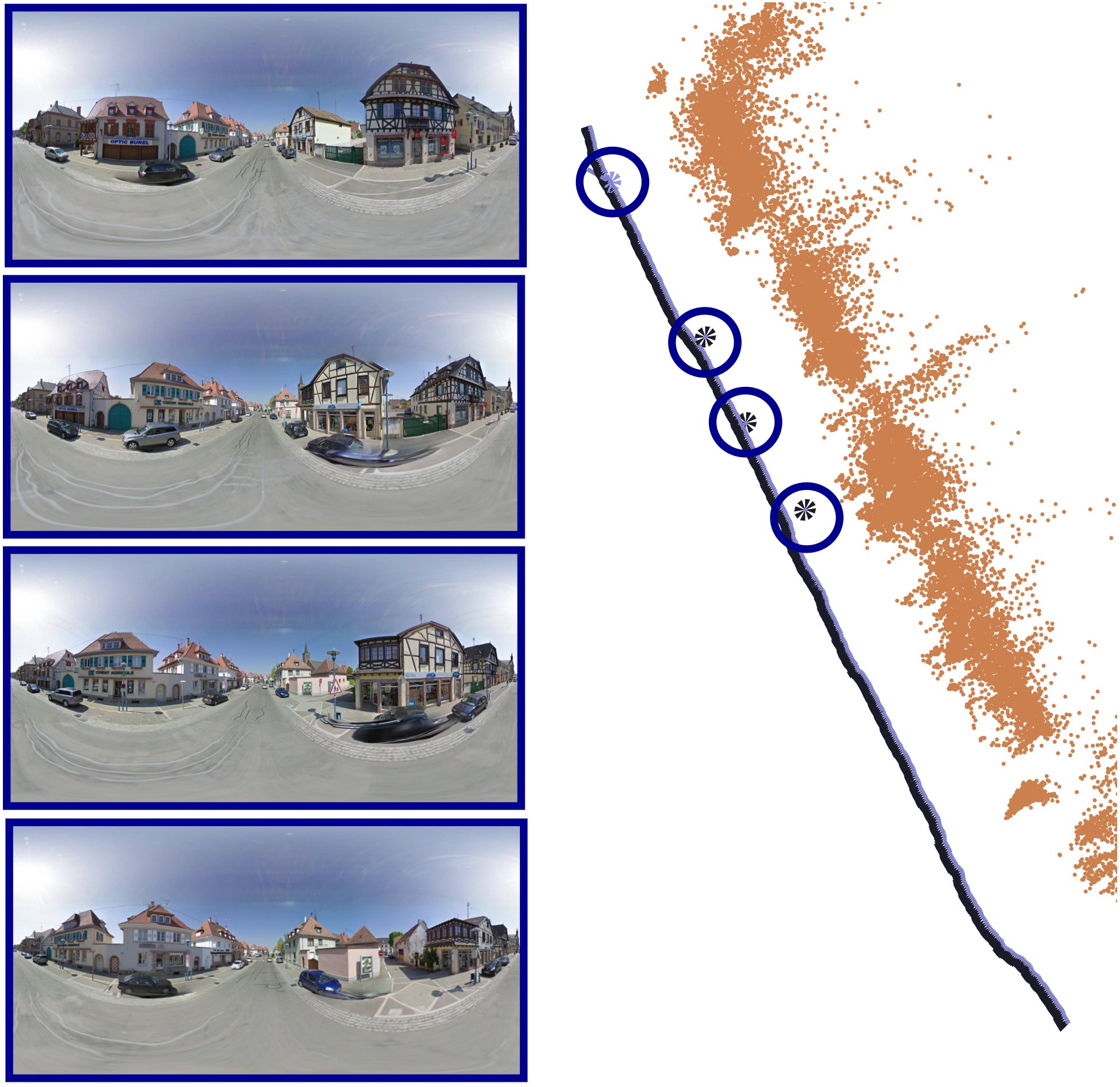}}
%\begin{tabular}{cc}
\subfigure{\includegraphics[width=0.22\textwidth]{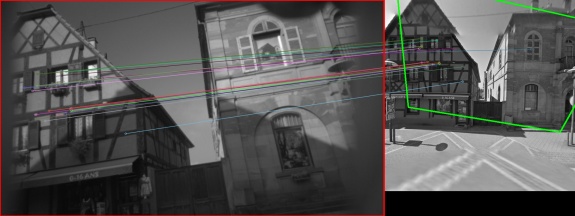}}
\subfigure{\includegraphics[width=0.22\textwidth]{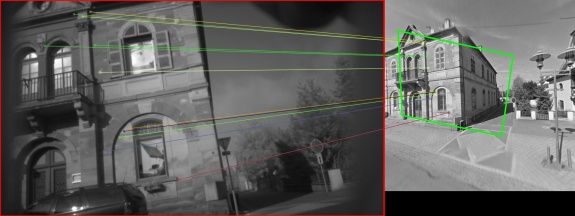}}
%\end{tabular}
\caption{Localization of a moving camera from Street View panoramas. The
  location of the panoramas are 27 Rue
  du Mal Foch, Marckolsheim, France. Four panoramas shown in
    top-left are localized with respect to the camera trajectory (black) and
    estimated 3D points (orange). The
    bottom two images show feature matching between Tango images and
    rectilinear views of the panorama.}
\label{covergirl}
\end{figure}

We formulate our approach as a non-linear least squares problem of two
objectives.  In the first, we estimate the 3D position of the points in the
environment from a short monocular camera trajectory. The short trajectories
are motivated by limiting the computation time, restricting the estimation
problem and the presence of abundant panoramic imagery. In the second, we find
panoramas that match the images and compute their 6DOF transformation with
respect to the camera trajectory and the estimated 3D points.  As the GPS
coordinates of the panoramic images are known, we obtain estimates of the
camera positions relative to the global GPS coordinates.  Our aim is not to
accurately model the environment or to compute loop closures for improving
reconstruction.  Our approach can be considered as a complement of GPS systems,
which computes accurate positioning from Street View panoramas. For this
reason, we tested our method on a Google Tango smartphone in two kinds of urban
environments, a suburban neighborhood in Germany and a main road of a village
in France.  Additionally, we quantify the accuracy of our technique by running
experiments in a large parking lot with ground truth computed from visual
fiducials. In the experiments, we show that with our technique we are able to
obtain submeter accuracy and robustly localize users or robots in urban
environments.

\section{Related work}

\label{ch:camera_loc_related} There exist previous literature about using
Street View imagery in the context of robotics and computer vision.
\citet{majdik13iros} use Street View images to localize a Micro Aerial Vehicle
by matching images acquired from air to Street View images.  Their key
contribution is matching images with strong view point changes by generating
virtual affine views.  Their method only solves a place recognition problem.
We, on the other hand, compute a full 6DOF metrical localization on the
panoramic images. In~\cite{majdikicra14}, they extended that work by adding 3D
models of buildings as input to improve localization.  Other researchers have
matched Street View panoramas by matching descriptors computed directly on
it~\cite{torii2011visual}.  They learn a distinctive bag-of-word model and use
multiple panoramas to match the queried image.  Those methods provide only
topological localization via image matching.  Related to this work is the topic
of visual localization, which has a long history in computer vision and
robotics, see~\cite{fuentes2012} for a recent survey. Various approaches have
been proposed to localize moving cameras or robots using visual
inputs~\cite{cummins-fabmap2_09,davison2007monoslam,konolige08tro_frameslam,
klingner13iccv,torii09iccv}. Our work is complimentary to such place
recognition algorithms as these may serve as a starting point for our method.
Topological localization or place recognition serves as a pre-processing step
in our pipeline. We use a naive bag-of-words based approach, which we found to
be sufficient for place recognition. Any of the above-mentioned methods can be
used instead to make the place recognition more robust.

Authors have also looked into localizing images in large scale metrical maps
built from structure-from-motion. \citet{irschara09cvpr} build accurate point
clouds using structure from motion and then compute the camera coordinates of
the query image. In addition, they generate synthetic views from the dense
point cloud to improve image registration. \citet{zhang06} triangulate the
position of the query image by matching features with two or more geotagged
images from a large database.  The accuracy of their method depends on the
density of the tagged database images. \citet{sattler12bmvc} also localize
query images in a 3D point cloud.  Instead of using all the matching
descriptors, they use a voting mechanism to detect robust matches. The voting
scheme enables them to select 3D points which have support from many database
images. This approach is further improved in ~\cite{sattler12eccv} by
performing a search around matched 2D image features to 3D map features and
vice versa. \citet{zamir10eccv} build a dense map from 100,000 Google street
view images and then localize query images by a GPS-tag-based pruning method.
They   provide a reliability score of the results by evaluating the kurtosis of
the voting based matching function. In addition to localizing single images,
they can also localize a non-sequential group of images.

Unlike others, our approach does not rely on accurate maps built with
a large amount of overlapping geotagged images.  As demonstrated by
the experiments, our approach requires only a few panoramas for
reliable metric localization with submeter accuracy.

\section{Method}
\label{ch:camera_loc_method}

\begin{figure}
\centering
\includegraphics[width=0.45\textwidth]{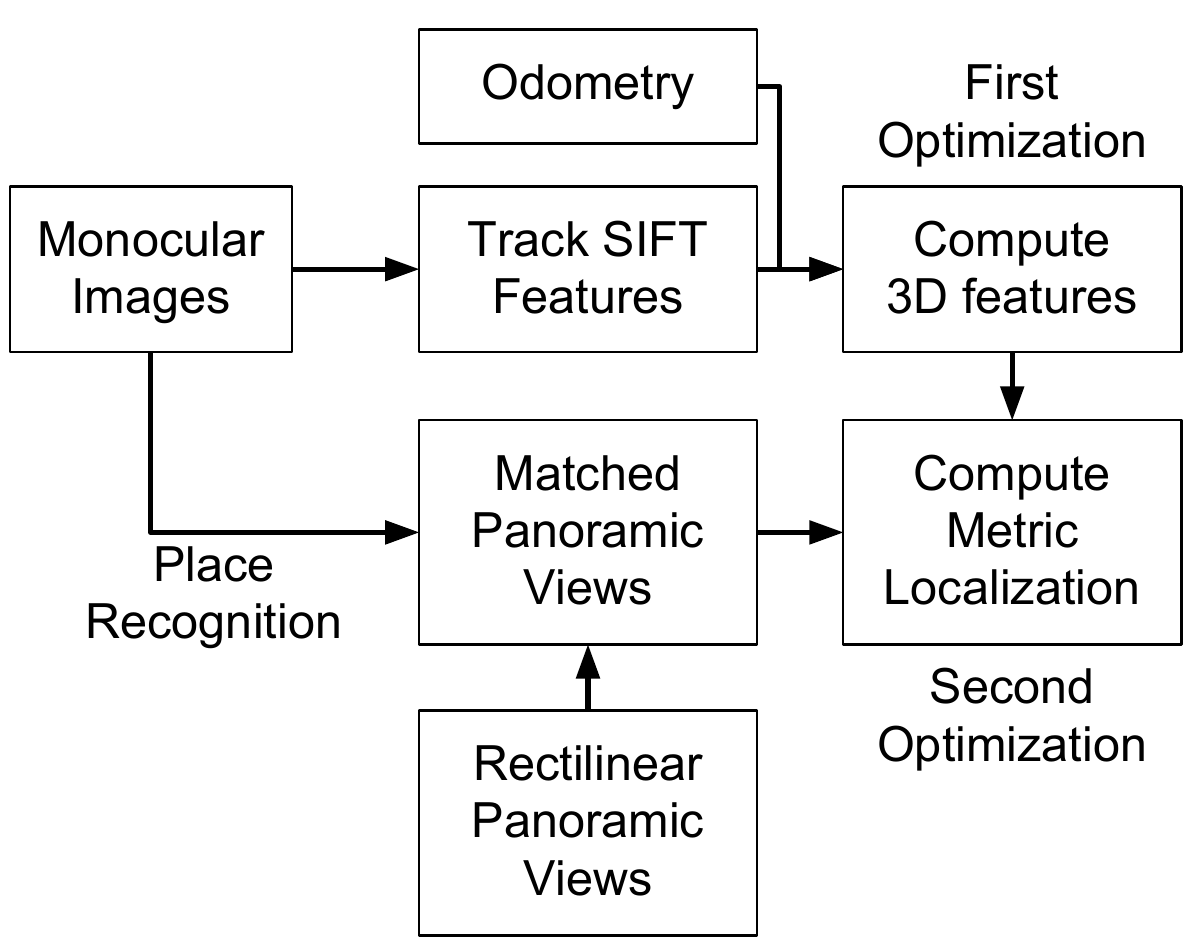}
\caption{Flowchart illustrating various modules in our pipeline.}
\label{fig:flowchart}
\end{figure}

In this section we outline the technical details for using Google's Street View
geotagged imagery as our map source for robot localization. Our goal is not to
built large scale accurate maps.  Instead, we want to approximately estimate the
3D position of the features relative to the camera positions and then compute
the rigid body transformation between the Street View panoramas and the
estimated points. This allows us to compute the GPS positions of the camera
position in global GPS coordinates. Our current implementation works offline.
The flowchart shown in \figref{fig:flowchart} illustrates the workflow between
the various modules.

\subsection{Tracking Features in an Image Stream}
\label{sec:tracking}

The input of our method is an image stream acquired from a monocular camera.  We
define $\seqset=(s_1, \ldots, s_\seqsize)$ as a sequence of $S$ images. A
sequence is implemented as a short queue that consists only of the last few
hundreds frames acquired by the camera.  An image $s_i$ is a 2D projection of
the visible 3D world, through the lens, on a camera's CCD sensor.  For
estimating the 3D position of the image points, we need to
collect bearing observations from several positions as the camera moves.

We take a sparse features approach for tracking features in the stream of camera
images. For each image $s_i$, we extract a set of keypoints computed by using
state-of-the-art robust feature detectors, such as SIFT \cite{lowe04ijcv}. A
description $d \in \mathcal{D}_i$ is computed from the image patch around each
keypoint.  $\mathcal{F}_i$ is the set of keypoints and descriptors and is
denoted as the feature set.

Each time a new image arrives, we find feature correspondences between $s_i$ and
$s_{i-1}$. We compute neighbor matches using
FLANN~\cite{muja12crv} between all elements of $\mathcal{D}_i$ and
$\mathcal{D}_{i-1}$. A match is considered valid if the distance to the best
match is $0.7$ times closer than the second best~\cite{lowe04ijcv}. As these
correspondences only consider closeness in descriptor space, in addition we employ a
homography constraint to consider the keypoint arrangement between two
images.  We use the keypoints of the matched descriptors for a RANSAC procedure
that computes the inlier set for the perspective transformation between the two
images.  We call a track $\mathcal{T}_j$, the collection of all the matched
keypoints relative to the same descriptor over the consecutive image frames
$\mathcal{S}$.  A track is terminated as soon as the feature cannot be matched
in the current image.  For an image stream $\mathcal{S}$, we collect the set of
tracks $\mathcal{T}_{\mathcal{S}}$ consisting of the features
$\mathcal{F}_{\mathcal{S}}$.

Note that tracks have different length. Some keypoints are seen from many
views, while others are seen from few. Intuitively, long tracks are good
candidates for accurate 3D point estimation as they have longer baselines.  We
only perform feature matching across sequential image frames. No effort is spent
on matching images which are not sequential: this work does not make any
assumption on the motion of the robot, on the visibility of the environment or
on the existence of possible loops.

\subsection{Non-Linear Least Squares Optimization for 3D Point Estimation}

\begin{figure}
\centering
\begin{tabular}{cc}
\includegraphics[width=0.22\textwidth]{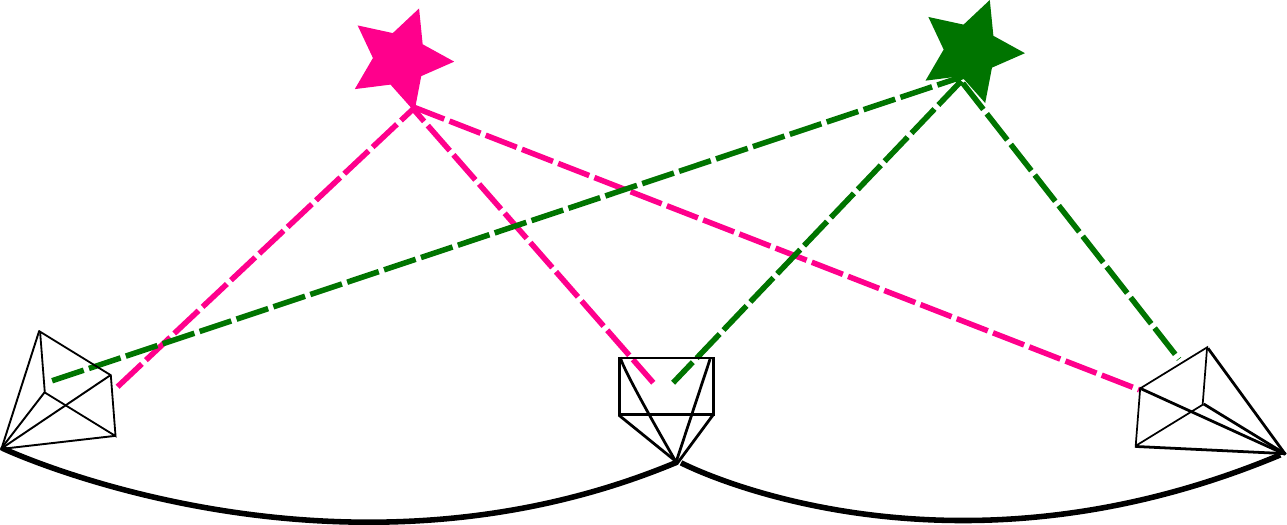} &
\includegraphics[width=0.22\textwidth]{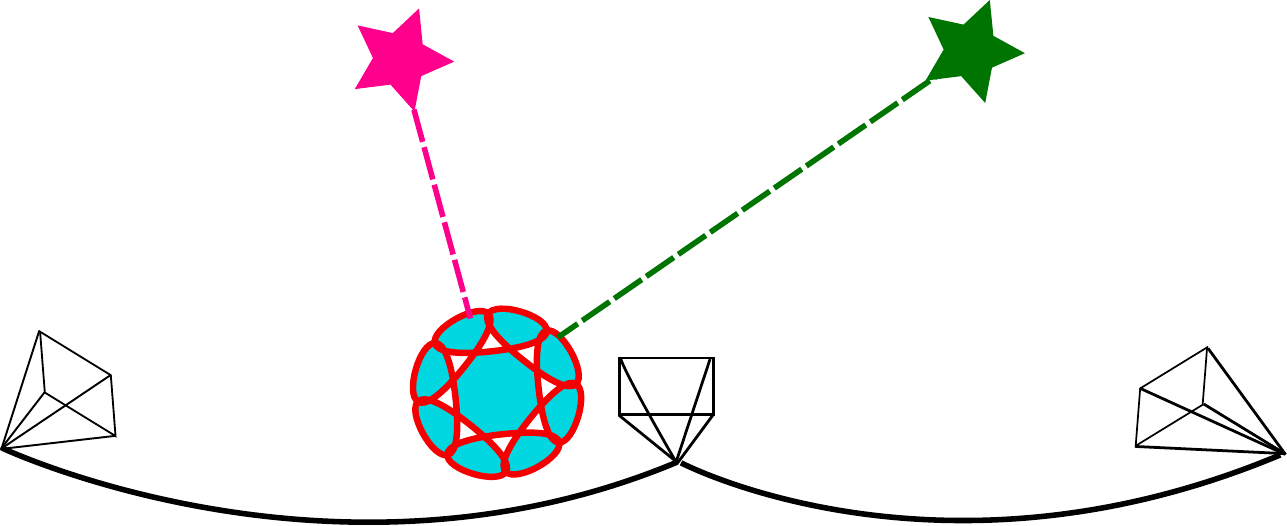} \\
\small{First Optimization} & \small{Second Optimization}\\
\end{tabular}
\caption{
Optimization problem for estimating the position of the features, $\by$ shown as
stars, and the camera positions $\bx$ shown as frustums. The dotted lines
represent bearing constraints while the solid black line represents the odometry
constraint. The right image shows the optimization problem for computing the
position of the panorama $\bp$ from the computed 3D points.}
\end{figure}

\label{sec:ba} The next step is to compute 3D points from the tracks
$\mathcal{T}_{\mathcal{S}}$.  In our system, we have rigid body odometric
constraints between consecutive camera poses $\bx_i$ and $\bx_{i+1}$, associated
to frame $s_i$ and $s_{i+1}$.  Our method is agnostic to the kind of odometry:
it can be computed by integrating IMUs, wheel encoders, or by employing an
IMU-assisted visual odometry. In our problem formulation, we consider the
monocular camera calibrated and all the intrinsic parameters known.

Each keypoint  in track $\mathcal{T}_j$ corresponds to a 3D point $\by_j$
observed in one of the images with pixel coordinates $u$, $v$.
If we consider a pinhole camera model, the camera matrix $\mathbf{C}$ projects
a point $\by_j$ into the camera frame:
\begin{eqnarray}
\mathbf{C} &=&   \left[ \begin{array}{ccc}
f_x & 0 &c_x   \\
0 & f_y & c_y   \\
0 & 0 & 1   \end{array} \right]
 \end{eqnarray}
The direction vector
\begin{eqnarray}
  \bd &=&\mathbf{C}^{-1} [u,v,1]^T,
\end{eqnarray}
can be interpreted as the direction of $\by_j$ with respect to the
camera center. Then, we compute the elevation and bearing angles:
\begin{eqnarray}
\theta &=& \operatorname{arccos}\left(\frac{d_z}{\sqrt{d^2_x + d^2_y +
  d^2_z}}\right) \label{eq:theta}\\
\varphi &=& \operatorname{arctan}\left(\frac{d_y}{d_x}\right) \label{eq:phi}
\end{eqnarray}
A least squares minimization problem can be described by the following equation:
\begin{eqnarray}
\bF(\bx,\by)&=& \sum_{ij}
\be_{ij}(\bx, \by)^T \bSigma_{ij} \be_{ij}(\bx, \by)   \nonumber \\
&+& \sum_{k} \be_{k,k+1}(\bx )^T \mathbf{\Lambda}_{k} \be_{k,k+1}(\bx)
\label{eq:optimization}
\end{eqnarray}

Here
\begin{itemize}
  \item  $\bx=(\bx_1^T,\;\ldots\;,\bx_n^T)^T$ is a vector of
    monocular camera poses, where each $\bx_i$ represents a 6DOF pose.
  \item $\by=(\by_1^T,\;\ldots\;,\by_m^T)^T$ is a vector of 3D points in the world associated to the tracked features.
  \item $\be_{ij}(\bx, \by)$ is a vector error function that computes
    the distance between a measurement prediction $\bzh_{ij}(\bx, \by)$ and a real measurement $\bz_{ij} =
    [\theta_{ij}, \phi_{ij}]$.  The error is $\bZero$ if $\bz_{ij} = \hat
    \bz_{ij}$, that is when the measurement predicted via
    $\bzh_{ij}(\bx, \by)$ from the states $\bx_i$ and $\by_j$ is equal
    to the real measurement.
  \item $\bzh_{ij}(\bx, \by)$ computes the bearing and azimuthal angles from camera pose $\bx_i$ to feature $\by_j$ in the camera frame.
    \item $\bSigma_{ij}$ represents the information matrix of a measurement
    that depends on the state variables in $\bx$.
  \item $\be_{k,k+1}(\bx)$ is a vector error from the predicted odometry measurements.
  \item $\mathbf{\Lambda}_{k}$ represent  the information matrix of the odometry.
    \end{itemize}

We initialize the camera position $\bx$ with odometry and the feature positions
$\by$ by triangulation.  We employ the
optimization framework g2o~\cite{kuemmerle11icra} as our non-linear least
squares solver.  First, we solve \eqref{eq:optimization} by keeping $\bx$ fixed:

\begin{eqnarray}
 \by^* &=& \argmin_{\by} \bF(\bx, \by)
\end{eqnarray}

This results in an improved estimation of $\by$.  Second, we perform a full
joint optimization of all the estimated 3D points $\by$ and camera poses $\bx$.

\begin{eqnarray}
 (\bx^*,\by^*) &=& \argmin_{\bx, \by} \bF(\bx, \by)
 \label{eqn:opt2}
\end{eqnarray}

The use of RANSAC helps improve the feature correspondences but does not
guarantee an absence of outliers. Therefore, the robust methods developed in the
previous chapters are used to improve the robustness against such errors. We use
Dynamic Covariance Scaling kernel, a robust M-estimator to improve convergence
and to handle wrong data associations~\cite{agarwal14icra-a}.

Note that we are not aiming at an accurate reconstruction of the environment.
In our approach, we only perform data association between sequential images as
we do not compute loop closures or perform large baseline feature triangulation.
There may be situations where a track is broken due to occlusions or changes in
the viewpoint. We do not try to merge tracks in such scenarios. This is avoided
for the process to be less computationally demanding. Doing a full
bundle-adjustment will definitely help in a better reconstruction of the
environment but that is not the goal of our work.

% Given the estimate of keypoints, we now need to
% match Street View panoramas with the estimated points $Z$.

\subsection{Matching of Street View Panoramas with Camera Images}

% In this section we outline how one can go about acquiring and using Street View
% imagery as panoramas.

Google Street View can be considered as an online browsable dataset consisting
of billions of street-level panoramic images acquired all around the world
\cite{google-sv}.  It is of key importance that each image is geotagged with a
GPS position. This position is highly accurate and it is the result of a careful
optimization at global-scale by Google \cite{klingner13iccv}.  In particular,
Street View images are acquired by vehicles with a special apparatus consisting
of cameras mounted around a spherical mounting.  All camera images
are stitched together to form a spherical panoramic image represented via a
\textit{plate carr\'{e}e} projection.  This results in a high quality image
often exceeding 20M pixels resolution for each panorama.

\begin{figure}
\centering
\subfigure {
\includegraphics[width=0.465\textwidth]{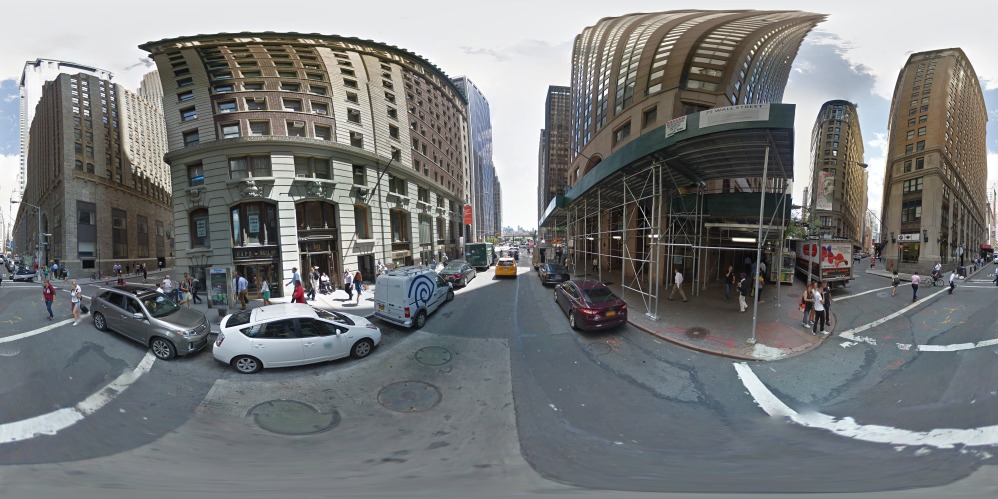}
}
\subfigure{
\setlength{\tabcolsep}{1pt}
  \begin{tabular}{cccccccc}
  \includegraphics[width=0.055\textwidth]{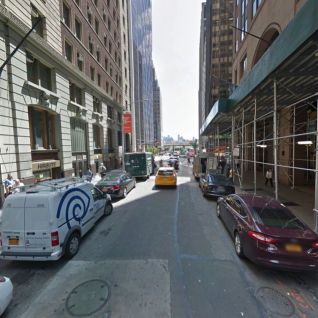} &
  \includegraphics[width=0.055\textwidth]{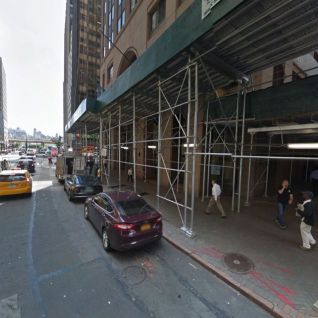} &
  \includegraphics[width=0.055\textwidth]{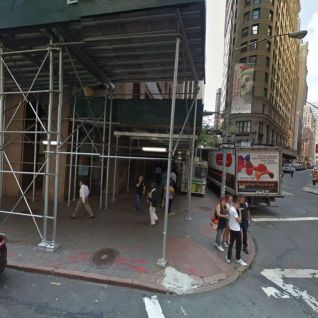} &
  \includegraphics[width=0.055\textwidth]{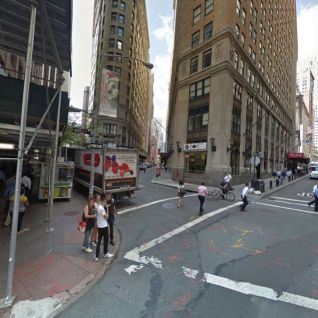} &
  \includegraphics[width=0.055\textwidth]{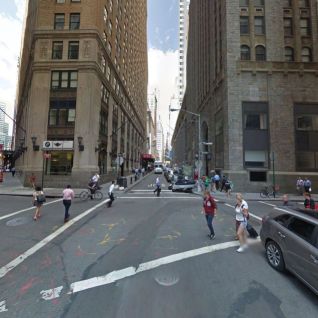} &
  \includegraphics[width=0.055\textwidth]{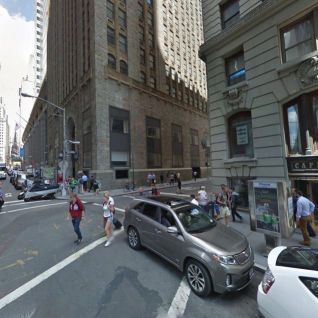} &
  \includegraphics[width=0.055\textwidth]{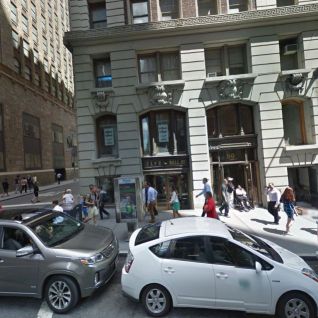} &
  \includegraphics[width=0.055\textwidth]{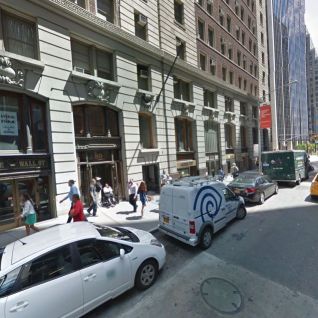} \\
  \end{tabular}
}
\caption{A panorama downloaded from Street View (top) with the extracted rectilinear
  views (bottom). Each image has a $90\degree$ field-of-view. These are
    considered pinhole cameras, free of distortion and they overlap horizontally to aid matching across image boundaries.
}
\label{fig:wall_st}
\end{figure}

Google provides public APIs for requesting virtual camera views of a given
panorama. These views are rectilinear projection of the spherical panorama with
a user selected field-of-view, orientation and elevation angle.  Rectilinear
views can be considered as undistorted images from a pinhole camera, free of
distortion.  A panorama can be selected via its GPS position or its ID.  An
example of a panorama acquired from Wall Street, New York, is illustrated in
\figref{fig:wall_st}. For
robustness, we extract rectilinear horizontal overlapping images. The
overlapping region aids in matching at image boundaries. We do not use the top
and the bottom image as it often contains only sky and floor.

% If prior information about the orientation of the monocular camera on the robot is known, a different point of projection can be
% generated from the panorama.
% The image looking above in the vertical direction may be
% interesting for narrow streets with tall buildings.
%
% \subsection{Bag of words based place recognition}
% In order to localize, we need to match a set of panorama to camera images.

In order to match panoramas with monocular camera trajectories we first need a
candidate set of panoramas. In our approach we rely on an inaccurate GPS sensor
to download all panoramic images in a certain large radius of approximately 1~km. The motivation
behind this approach is that a robot will roughly know which neighborhood or
city it is operating in. First, we collect all the rectilinear views from the
panoramic images $\mathcal{P}$ and build a bag-of-words image retrieval system
\cite{fei05cvpr}.  We compute SIFT keypoints and descriptors $\mathcal{F_P}$ for
all rectilinear panoramic views in $\mathcal{P}$ and group them with k-means
clustering to generate a visual codebook.  Once the clusters are computed and
we describe each image as histograms of visual words, we implement a TF-IDF
histogram reweighing.  For each camera image, we compute the top $K$ images from
the panoramic retrieval system, which have the highest cosine similarity.  This
match can be further improved by restricting the search within a small radius
around the current GPS location or from the approximate location received from
cellular network towers. Second, we run a homography-based feature matching,
similar to the one used for feature tracking in \secref{sec:tracking} to select
the matching images from $K$. These matched images are used as the final
candidate panoramic images used to compute the global metric localization
explained in the next section.

\subsection{Computing Global Metric Localization}

To localize in world reference frame, we compute the rigid body transformation
between the moving camera imagery and the geotagged rectilinear panoramic views.
We look for the subset of features $\mathcal{F}^* = \{\mathcal{F}_\mathcal{P}
\cap \mathcal{F}_\mathcal{S} \}$ that are common between the monocular images
$\mathcal{S}$ and the top $K$ panoramic views. The 3D positions of
$\mathcal{F}$ have been estimated using the methods in \secref{sec:ba}.  We
consider the rectilinear views as perfect pinhole cameras: the focal length
$f_x, f_y$ are computed from the known field-of-view; $c_x,c_y$ is assumed to be
the image center.  We follow the same procedure of \secref{sec:ba} for computing
the azimuthal and bearing angles for each element of $\mathcal{F}^*$ using
\eqref{eq:theta} and \eqref{eq:phi}.

To localize the positions of the $K$ panoramas from the feature positions $\by$,
we formulate another non-linear least squares problem similar to
\eqref{eq:optimization}:
\begin{eqnarray}
\bG_1(\bp, \by )&=& \sum_{ij}
\be_{ij}(\bp, \by)^T \bSigma_{ij} \be_{ij}(\bp, \by)
\label{eq:pano_unjoined}
\end{eqnarray}
where
\begin{itemize}
  \item  $\bp=(\bp_1^T,\;\ldots\;,\bp_{8 \times m}^T)^T$
  is a 6DOF vector of poses associated to the rectilinear views taken from $m$ panorama images.
  \item $\by$ is the vector of the estimated 3D points.
  \item $\be_{ij}(\bp, \by)$ is the same error function defined for the
  optimization \eqref{eq:optimization}. This is computed for all $\mathcal{F}^*$
  between panorama view $\bp_i$ and 3D points $\by_j$.
 %   \item $\be_{k,k+1}(\bx)$ is a vector error from the predicted odometry measurements.
  \item $\bSigma_{ij}$ represents the information matrix of the measurement.
\end{itemize}
For robustness, we connect multiple views from the same panorama that
are constrained to have the same position but a relative yaw offset of
$90\degree$.

The optimization problem becomes
\begin{eqnarray}
\bG_2(\bp, \by )&=& \bG_1(\bp, \by )  \nonumber\\
&+& \sum_{k} \be_{k,k+1}(\bp )^T \mathbf{\Lambda}_{k} \be_{k,k+1}(\bp)
\label{eq:pano_joined}
\end{eqnarray}
where $\be_{k,k+1}(\bp)$ is the error between two rectilinear
views computed from the same panorama. The optimal value for $\bp*$
can be found by solving:

\begin{eqnarray}
 \bp^* &=& \argmin_{\bp} \bG_1(\bp, \by)
  \label{eq:opt_unjoined}
\end{eqnarray}
or alternatively by solving:
\begin{eqnarray}
 \bp^* &=& \argmin_{\bp} \bG_2 (\bp, \by)
  \label{eq:opt_joined}
\end{eqnarray}

After optimization, the panoramic views are in the frame of reference
of the monocular camera trajectory $\bx$. Now, it is trivial to
compute the relative offset between the map and the panorama, hence
computing precise global GPS coordinates of the camera images.
\footnote{In our experiments, some of the panoramas were manually
acquired with a cell phone and hence the panorama rig is not fixed. By
optimizing the additional rig parameters we are more robust to small errors in
the panorama building process. Additionally, we do not have any constraints between
different panoramas collected from different places. Each
panorama is independently optimized.}

\section{Experimental Evaluation}
\label{ch:camera_loc_evals}
We evaluated our method in two different scenarios. In the first, we considered
an outdoor parking lot area and placed visual fiducials for estimating the
accurate ground truth. In the
second, we used a Google Tango device in two different urban scenarios.  The
first scenario is in Freiburg, Germany where we personally uploaded panoramas
acquired with mobile devices. This is required as Street View is only partially
available in Germany. For the second scenario, we tested our technique on
panoramas from Street View collected by Google in Marckolsheim, France.  All of
the panoramas used in these experiments are publicly available.

\begin{figure}
\centering
\subfigure{
\includegraphics[height=.16\textwidth]{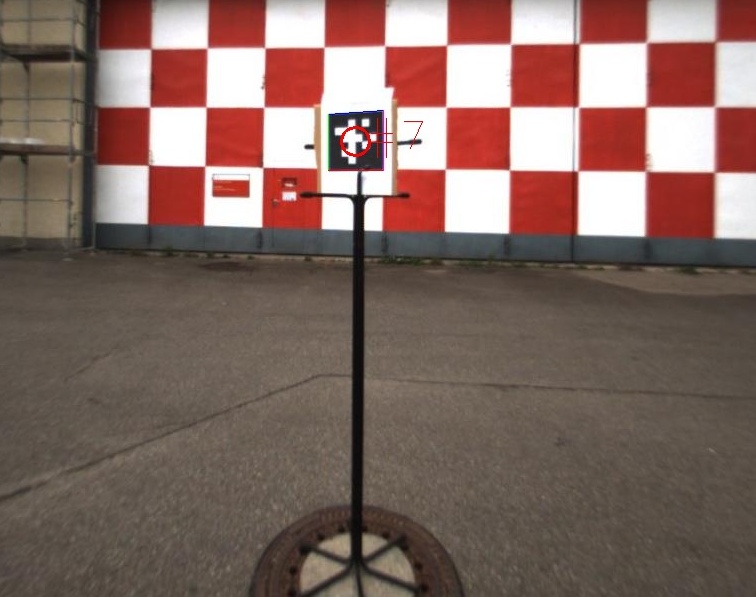}}
\subfigure{
\includegraphics[height=.16\textwidth]{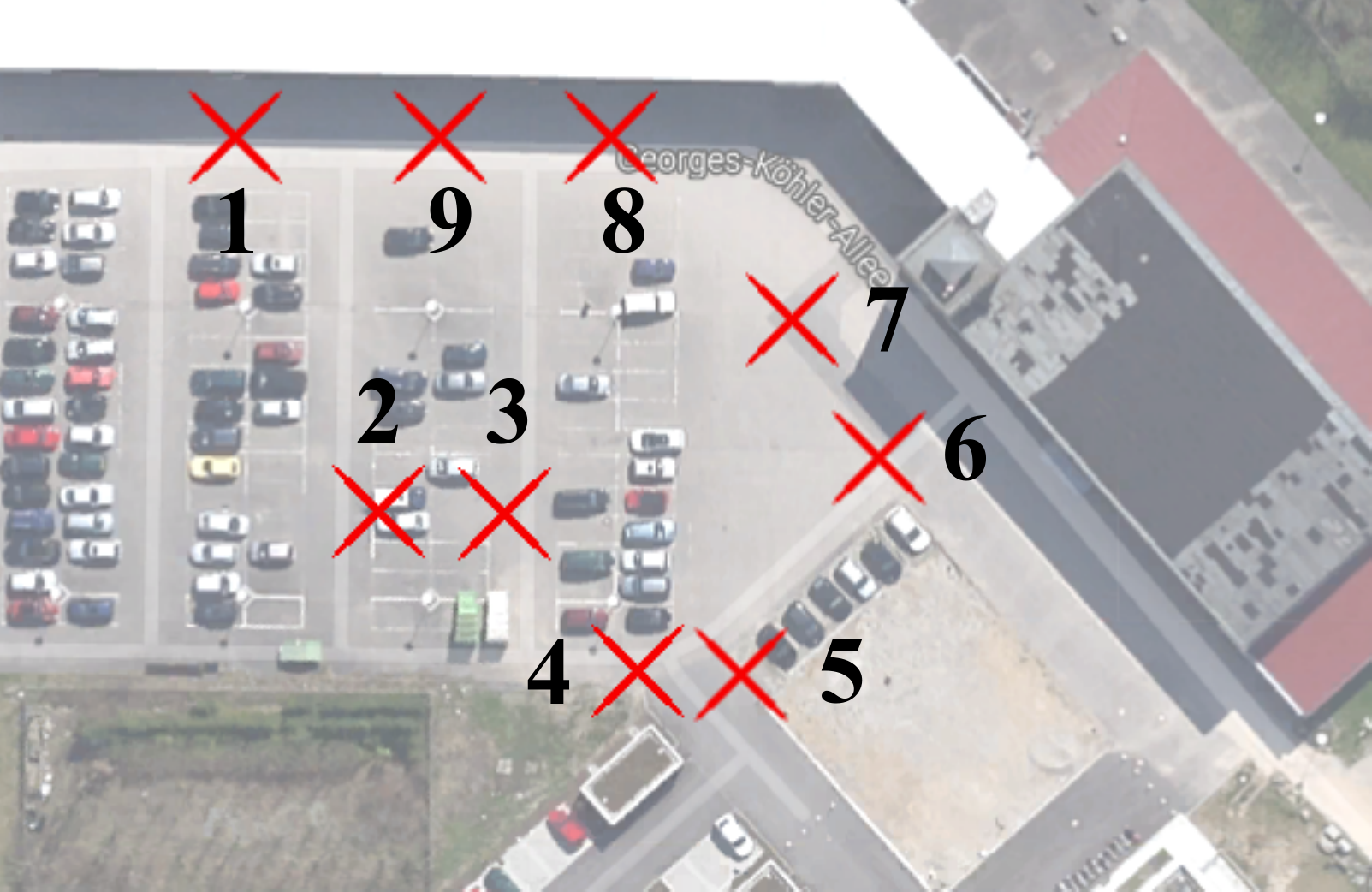}}
  \caption{The left figure shows an example
    of an AprilTag placed above a manhole from where a panoramic image
    was acquired. The right figure illustrates an aerial view of the parking lot for the parking lot
    experiment.  Red crosses highlighting the positions of the
    panoramas. The numbers represent the AprilTag ID. }
\label{fig:parking_lot}
\end{figure}

\subsection{Metric Accuracy Quantification}

The parking lot experiment is designed to evaluate the accuracy of our method.
It is full of dynamic objects and visual aliasing. Additionally, most of the
structures and buildings are only on the far-away perimeter of the parking lot.

\begin{figure}
\centering
\subfigure{\includegraphics[height=.18\textwidth]{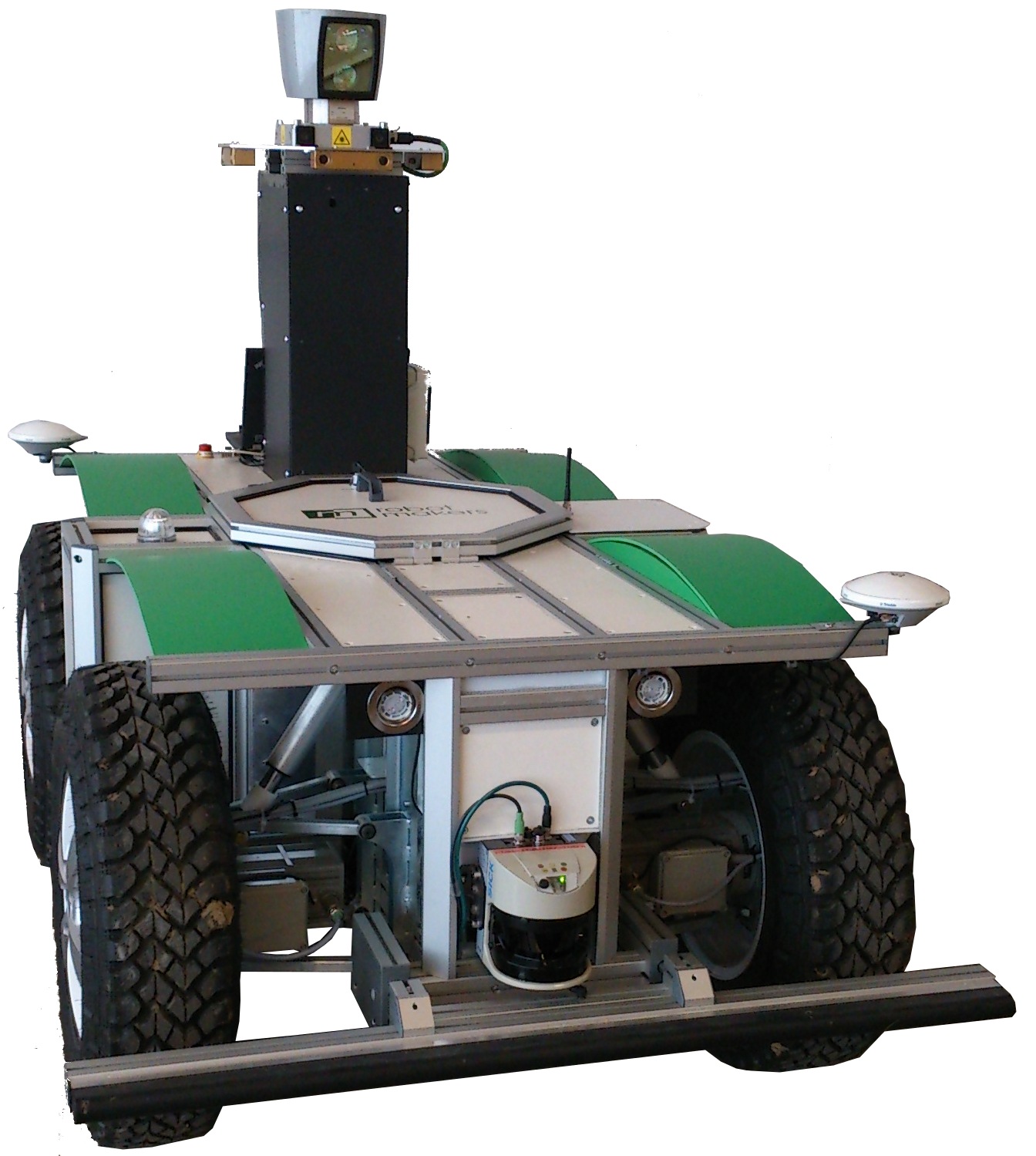}}
\subfigure{\includegraphics[height=.18\textwidth]{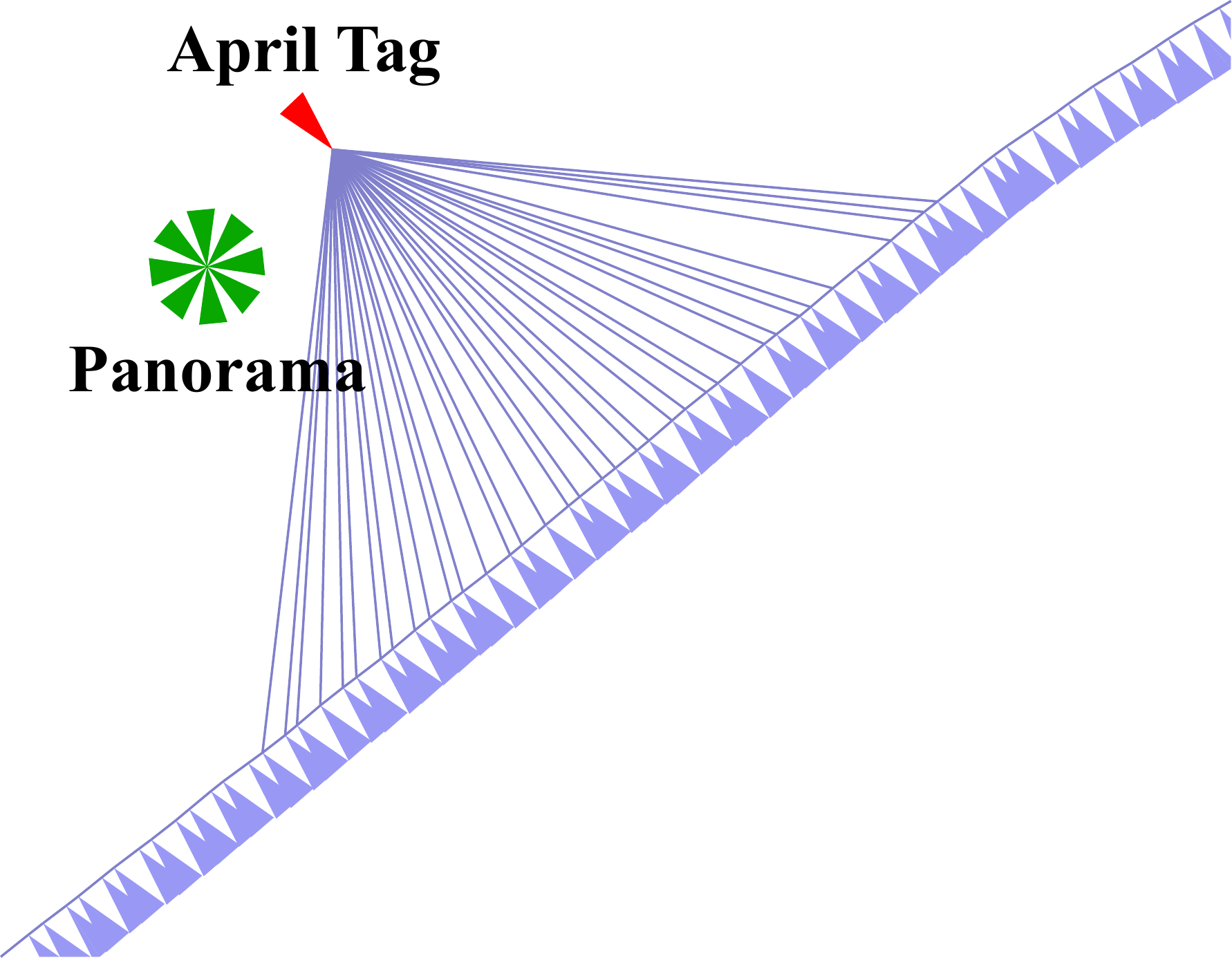}}
\caption{The left figure shows the robot used to conduct the parking lot
experiments and the right figure illustrates the
final monocular camera positions with the estimated position of the panorama and
April tag in the parking lot.}
\label{fig:tango_ball}
\end{figure}

Using GPS as ground truth is not sufficient as our method aims at providing
accurate estimations, potentially better than GPS accuracy.  For reference, we
collected spherical panoramas by using a smartphone, on visually distinct
landmarks such as manholes.  Then, as a ground truth we placed visual fiducials,
namely AprilTags~\cite{olson11icra} above the manholes from where the panoramas
were acquired. The fiducials serve as a way to compute the ground truth
positions of the manually acquired panoramas.   AprilTags come with a robust
detector and allow for precise 3D positioning.  We use the tag family 36h11 and
the open source implementation available from ~\cite{kaess-apriltag}.
\figref{fig:parking_lot} shows one such tag from the view of the camera with the
tag detection and detected id superimposed on the image.
\figref{fig:parking_lot} also illustrates the aerial view of parking lot with
the position from where the panoramas were generated (red crosses).  The numbers
represent IDs for each April Tag.  To have a fine estimate of the panoramic
image pose, we use non-linear least squares to optimize for the full 6D tags
positions from the computed camera poses as illustrated in
\figref{fig:tango_ball}.

\begin{figure*}[t!]
\centering
\subfigure[$135$]{\includegraphics[width=0.3\textwidth]{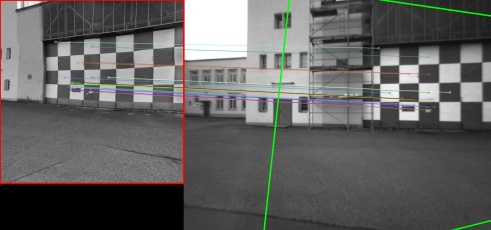}
\label{fig:matchiwthpano}
}
\subfigure[$180$]{\includegraphics[width=0.3\textwidth]{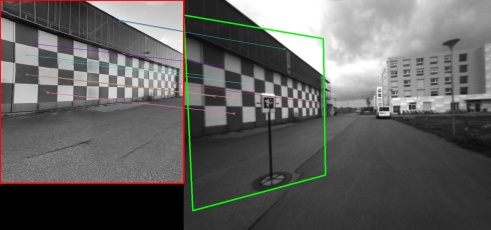}
\label{fig:matchiwthpano2}
}
\subfigure[$225$]{\includegraphics[width=0.3\textwidth]{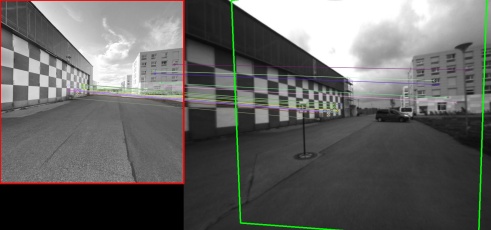}
\label{fig:matchiwthpano3}
}
\caption{Matching 3 views of the same panorama to monocular images.
  $90\degree$ field-of-view rectilinear projections and the corresponding feature
    matches for each view can be seen. Homography projection of the panoramas on
    the monocular image is shown in green.}
\label{fig:pano_match}
\end{figure*}

For these experiments, we used a robot equipped with an odometry estimation
system and a monocular 100$\degree$ wide field-of-view camera. We performed $4$
runs around all $9$ AprilTags and $1$ shorter run. In total we performed a total
of $5$ different runs in the parking lot.  The position of panoramas and
AprilTags are computed with respect to the camera positions.
Tables~\ref{tab:unjoined} and~\ref{tab:joined} report the error between the
computed pose of the panorama and the associated tag for the $5$ runs.
~\figref{fig:pano_match} shows examples of feature matches between three
panorama views and camera images. Each of the three images in
\figref{fig:pano_match} show the matched features and homography of the
rectified panorama projected on to the image acquired from the monocular camera.

\begin{table}
\centering
\begin{tabular}{ccccccccc}
\midrule
%6 & 11 & 9 & 13 & 5 & 7 & 14 & 10 & 8 \\
1  & 2  & 3 & 4  & 5 & 6 & 7  & 8  & 9 \\
\midrule
 6.00 & 3.14 & 0.99 & 1.65 & 5.36 & 2.94 & 1.22 & 0.53 & 1.29\\
 0.72 & 1.07 & - & 9.62 & 4.91 & 2.00 & 5.53 & 0.70 & 1.28\\
 - & - & - & - & - & 3.35 & 0.74 & 1.07 & 4.03\\
 - & - & 0.37 & 1.03 & 3.92 & 5.34 & 2.07 & 0.39 & 1.07\\
 - & 0.47 & 0.58 & 0.37 & 12.38 & 5.89 & 1.53 & 0.55 & 2.59\\

\bottomrule
\end{tabular}
\caption{
Error (in meters) between estimated pose of each individual panoramic view
compared to the ground truth tag.}
\label{tab:unjoined}
\end{table}

\begin{table}
\centering
\begin{tabular}{ccccccccc}
\midrule
%6 & 11 & 9 & 13 & 5 & 7 & 14 & 10 & 8 \\
1  & 2  & 3 & 4  & 5 & 6 & 7  & 8  & 9 \\
\midrule
 10.11 & 3.14 & 1.09 & 1.65 & 3.44 & 3.61 & 1.22 & 0.36 & 1.29\\
 0.72 & 1.07 & - & 5.80 & 3.27 & 0.80 & 1.51 & 0.68 & 1.28\\
 - & - & - & - & - & 3.35 & 0.70 & 1.07 & 0.90\\
 - & - & 0.37 & 1.03 & 2.84 & 5.34 & 2.07 & 0.60 & 1.07\\
 - & 0.47 & 0.59 & 0.42 & 3.75 & 5.88 & 1.53 & 0.55 & 3.39\\

\bottomrule
\end{tabular}
\caption{Error (in meters) between estimated pose of the connected panoramic views compared to the ground truth tag.}
\label{tab:joined}
\end{table}

\begin{figure}
\centering
\includegraphics[width=0.4\textwidth]{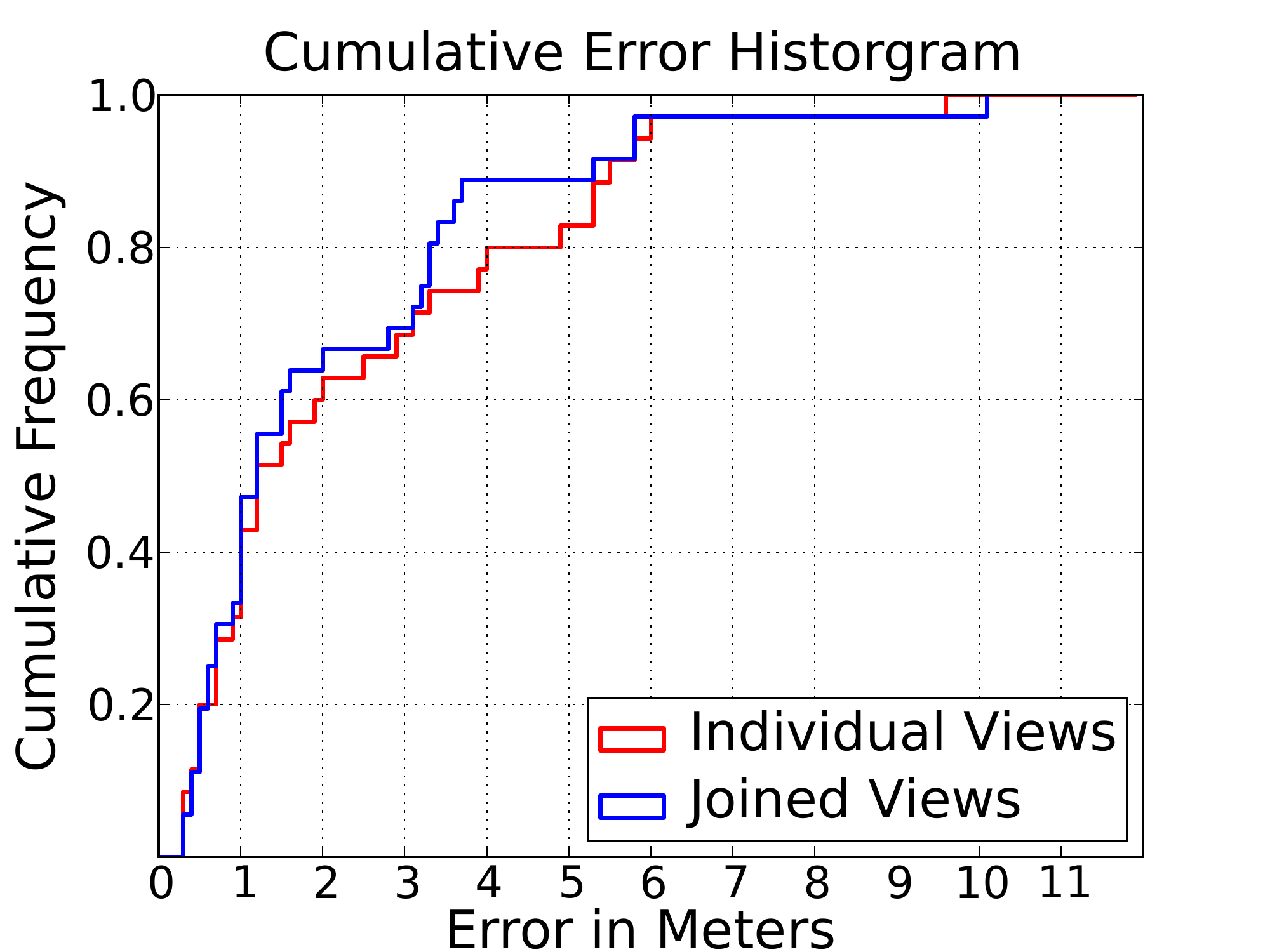}
\caption{Cumulative error histogram for the parking lot experiment. The optimization of the connected panoramic views (blue) improves the performance.}
\label{fig:graph}
\end{figure}

The errors reported in \tabref{tab:unjoined} correspond to optimizing the
individual views of the panoramas without any constraints among them. This
corresponds to the optimization in \eqref{eq:opt_unjoined}.
That is, if two views of a panorama match at a certain place, we optimize them independent of
each other. \tabref{tab:joined} reports errors when all the views of the
panoramas are connected together. This corresponds to the optimization in
\eqref{eq:opt_joined}.
Connecting views from the same panorama together improves the accuracy
as can be seen from \figref{fig:graph}.  The system does not report
localization results if the matching rectilinear views are estimated
too far with respect to the current pose.

The panorama acquired from the position of tag id 5 and 6 is
localized with least accuracy as most of the estimated 3D features are
far way ($>50m$). The panoramas acquired from the position of tags
8 and 9 are localized with the highest accuracy as the tracked
features are relatively closer ($15m-20m$).

As expected, the localization accuracy decreases with increase in the
distance to tracked features. Points that are far away from the camera
show small motion. In these cases, small errors in the odometry
estimate and in the keypoint position in the image cause considerable
errors in the estimated feature distances in 3D.  Nevertheless, about
40\% of the runs we are within a $1m$ accuracy, 60 \% within
$1.5m$. This is significantly lower than the accuracy on mobile
devices (5 to 8.5 m) which use cellular network and GPS~\cite{zandbergen2011positional}.
% or the standard GPS resolution of $6.7m$\xxx{cite from where you got this}.

Despite our efforts in providing accurate ground truth, this is not
free of errors. Especially because the exact center of the panorama is
unknown. Manually acquired panoramas are difficult to generate and
often the camera center moves. The individual images which are
stitched together often do not share the same exact camera center.

\begin{figure}
\centering
\begin{tabular}{cc}
\multicolumn{2}{c}{\includegraphics[width=0.45\textwidth]{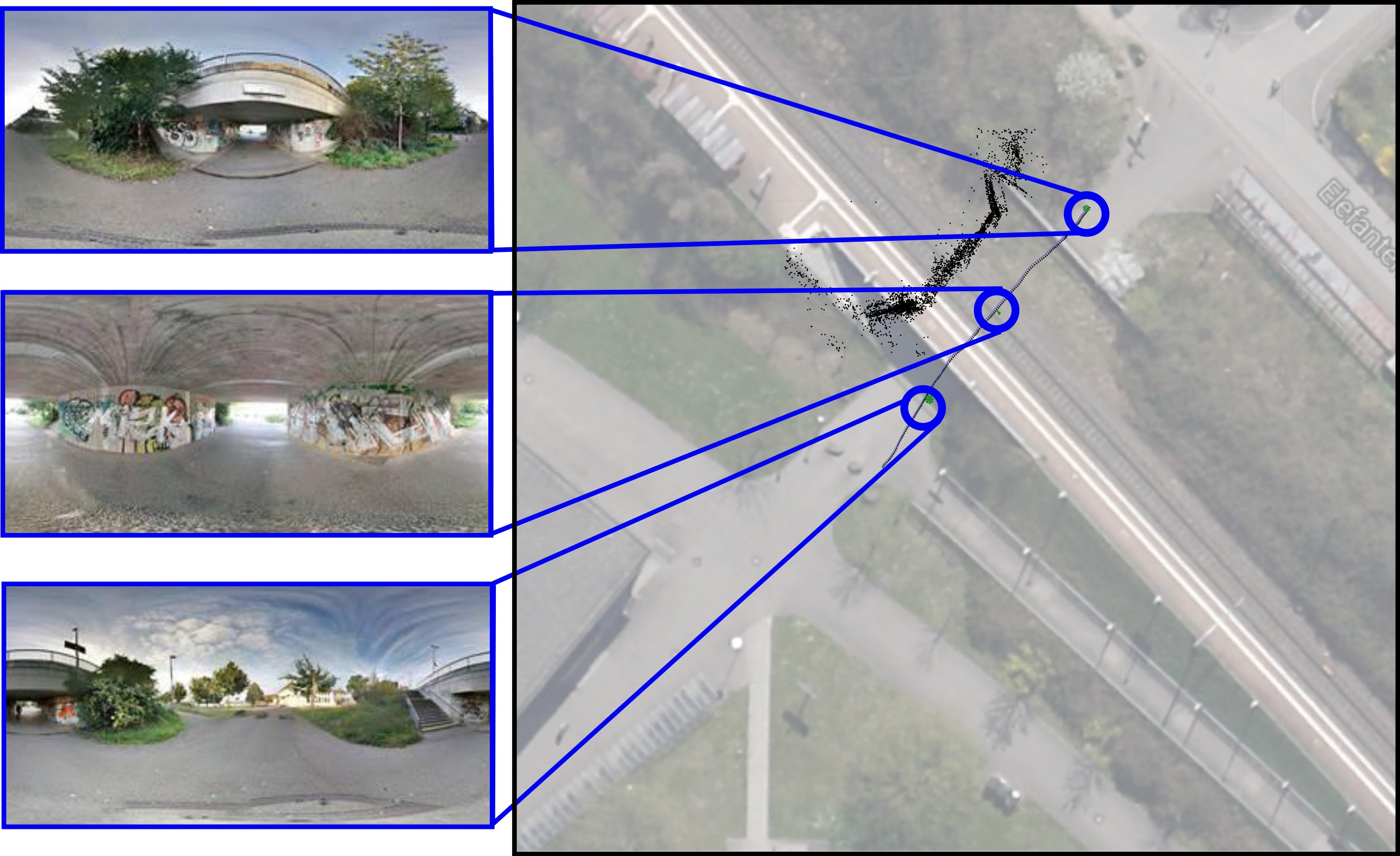}}
\\
\includegraphics[width=0.22\textwidth]{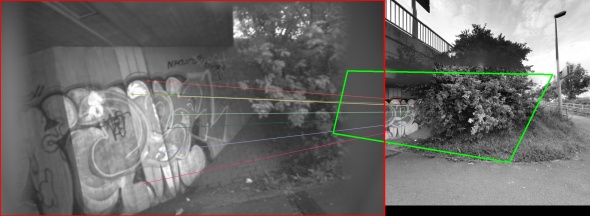} &
\includegraphics[width=0.22\textwidth]{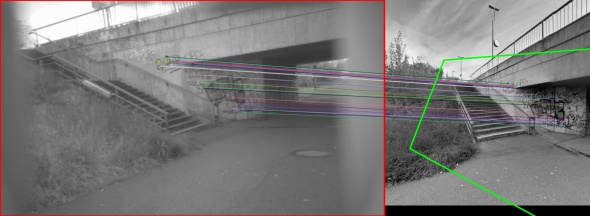}
\\
\includegraphics[width=0.22\textwidth]{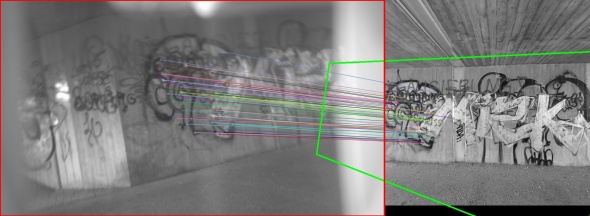}
                       &
\includegraphics[width=0.22\textwidth]{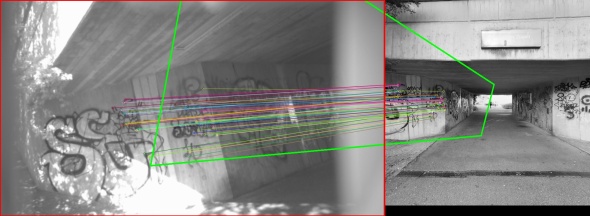} \\
\end{tabular}
\caption{Matches between monocular images and Street View
panoramas for a railway underpass used for the Tango
  experiments. The middle image shows the aerial imagery of the location,
superimposed with localization results of the panoramic
  view with respect to the camera trajectory.
The second panorama is acquired under the bridge while
  the other two are outside. The images on the right show example matches that
were found between the monocular images and extracted rectilinear panoramic
views.}
\label{fig:tango_bridge_pano}
\end{figure}

\subsection{Urban Localization with a Google Tango Device}

\begin{figure}
\centering
\includegraphics[height=0.16\textwidth]{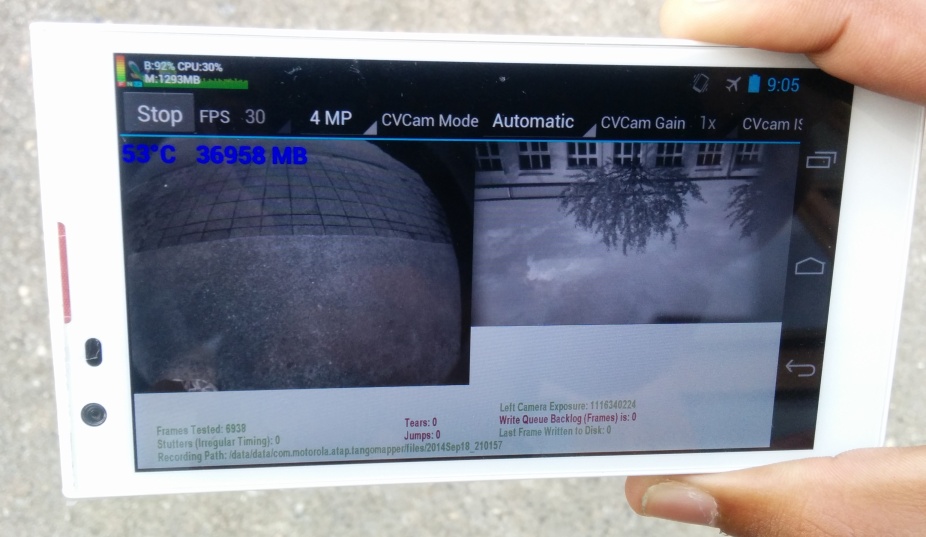}
%\quad
\includegraphics[height=0.16\textwidth]{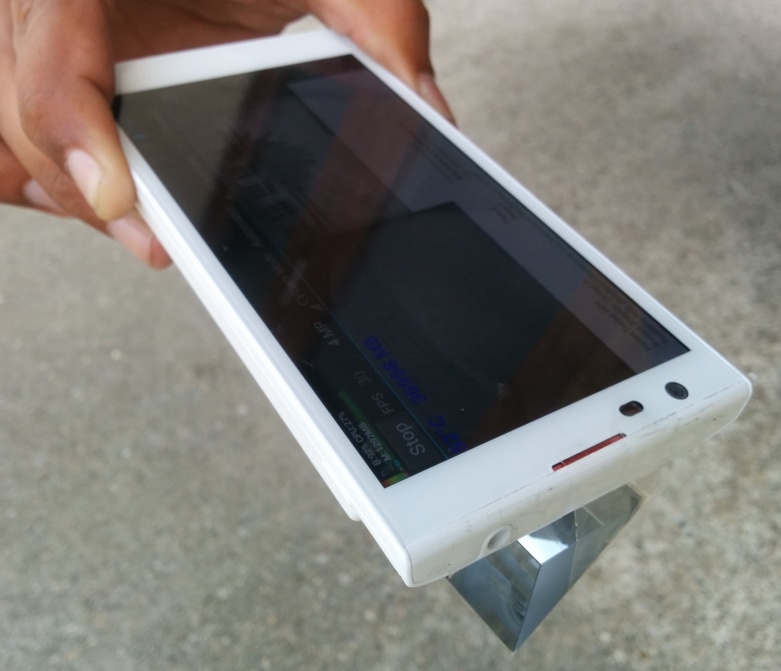}
\caption{Google Tango with the prism attached to the narrow field of view camera. The screen shows the camera used for visual odometry pointing downwards, while the
narrow field of view camera points sidewards.}
\label{fig:tango_prism}
\end{figure}

In order to show the flexibility of our approach, we evaluated our algorithm
with a Google Tango device in two urban environments.  We used the integrated
visual inertial odometry estimated on the Tango device for our method.  Tango
has two cameras: one that has a high resolution but a narrow field-of-view, and
another one, that has a lower resolution but a wider field-of-view.  The narrow
field-of-view camera has a frame rate of $5$Hz, the other streams at $30$Hz.  We
use the higher resolution camera as the monocular image source for our
framework, meanwhile the wide angle camera is used internally by the device for
the odometry estimates.  Throughout our experiments, we found the odometry from
Tango to be significantly more accurate indoor than outdoor. This is probably
due to a relatively weak feature stability outdoors and the presence of only
small baselines when navigating in wide areas.  To alleviate this problem, we
mounted a mirrored $45-90-45$ prism on the narrow-field-of-view camera and
pointed the wide field-of-view to the floor.  In this way, the Tango device
reliably tracks features on the asphalt and computes accurate odometry
estimates, meanwhile the other camera points at the side.
\figref{fig:tango_prism} shows the prism mounted Tango device.

\begin{figure}
\centering
\subfigure{\includegraphics[height=0.18\textwidth]{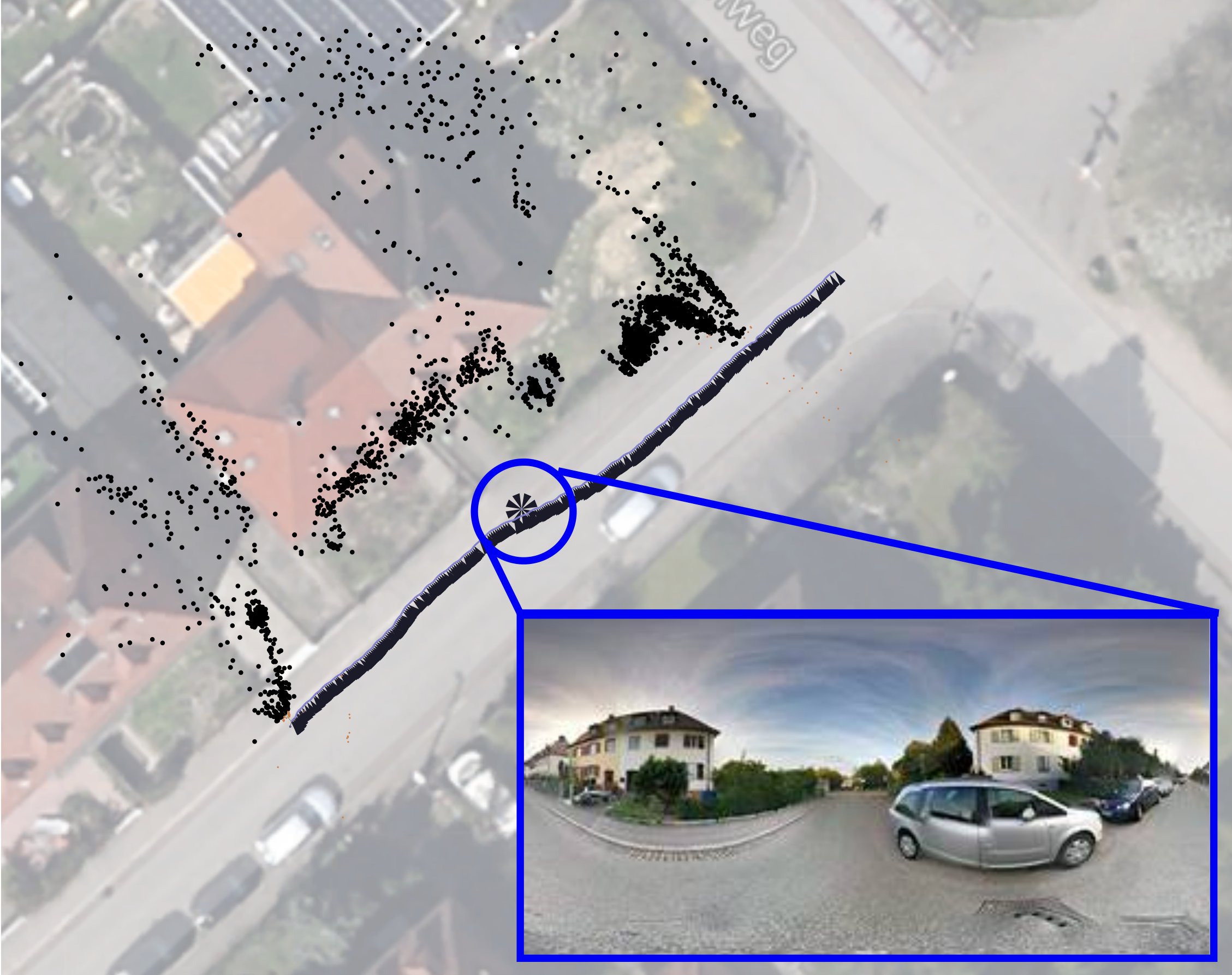}
}
\subfigure{\includegraphics[height=0.18\textwidth]{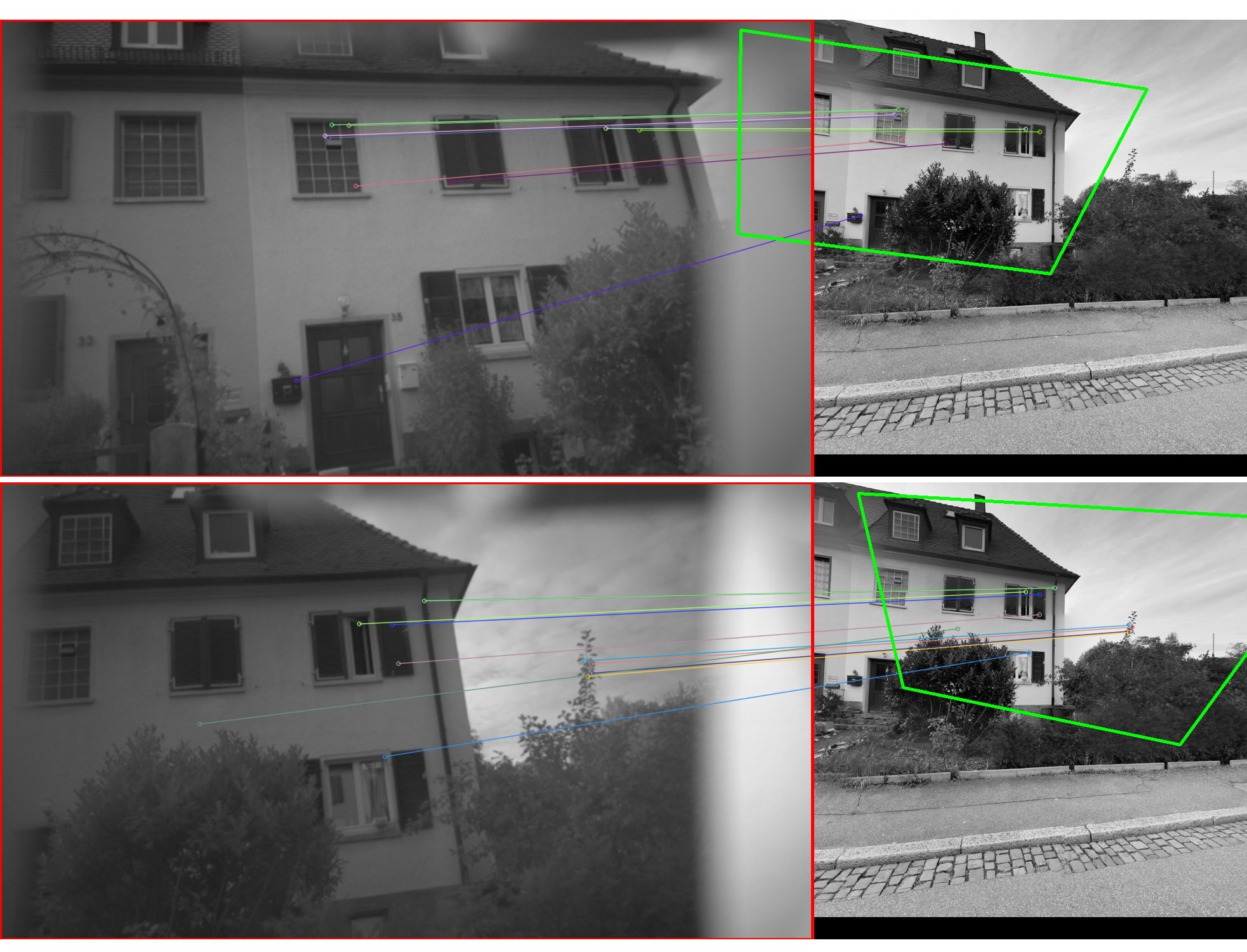}
}
\caption{Optimized 3D points with the estimated panorama position
  overlayed on Google maps (top). An example of matching between
  panorama views and Google Tango images. As both cameras are pointing in
different directions, the features used internally for visual inertial odometry are
different from the features used for localizing Street View panoramas.}
\label{fig:elephant1}
\end{figure}

%\begin{figure}
%\centering
%\includegraphics[width=0.45\textwidth]{pics/tango/bridge.pdf}
%\caption{Aerial imaginary of a railway underpass used for the Tango experiments.
%The middle panorama is acquired under the bridge while the  other two are outside.}
%\label{fig:tango_bridge_panos}
%\end{figure}

The first urban scenario has been run on roads around the University campus in
Freiburg, Germany.  In particular, around the area with GPS coordinates
$48.0125518,7.8322567$.  The panoramas used in the Tango experiments are public
and can be viewed on Street View.  In the experiment, we crossed a railway line
by using an underpass where GPS connection is lost, see
\figref{fig:tango_bridge_pano}.  Our method is able to estimate 3D points from
the images acquired from Tango and then match them to the nearby panoramic
images, see \figref{fig:tango_bridge_pano}.  Then, we moved into the suburban
road with houses on both sides. This location is challenging due to the fact
that all houses look similar.  Also in this case, our approach is able to
correctly estimate the 3D points of the track and localize the nearest panorama,
see \figref{fig:elephant1}.  In the figure, the black points are the estimated
3D points while the circles in the center of the image are the positions of the
panorama views. The pose of the Tango device is overlayed on the street.

To test our technique in a Street View panorama acquired by Google, we ran
another experiment on the main road of the village of Marckolsheim, France.
Despite being a busy road, our technique correctly estimated 3D points from the
Tango image stream and successfully estimated the panorama positions, see
\figref{covergirl}.

%\begin{figure}
%\centering
%\subfigure[]{\includegraphics[width=0.22\textwidth]{pics/tango/tango_bridge_0.pdf}}
%\subfigure[]{\includegraphics[width=0.22\textwidth]{pics/tango/tango_run1.pdf}}
%\caption{Localization of the panoramic view with respect to the camera trajectory.}
%  \label{fig:bridge}
%\end{figure}

%\begin{figure}
%\centering
%\begin{tabular}{cc}
%\includegraphics[width=0.22\textwidth]{pics/tango/scaled_france_good1.jpg} &
%\includegraphics[width=0.22\textwidth]{pics/tango/scaled_france_good2.jpg} \\
%\end{tabular}
%\caption{
%Matches between Tango images and Street View panoramas from Marckolsheim,
%        France.}
%\label{fig:tango_bridge_pano}
%\end{figure}
\section{Discussion}

Our method can use any kind of odometric input. In the case of Tango, the
odometry is based on the work of \citet{mourikis07icra} that makes use of
visual odometry and IMUs to generate accurate visual-inertial odometry (VIO).
This system is offered by the Google's Tango device libraries.  When
implemented on Tango, our method uses the two onboard cameras. One is the wide
angle camera, that is used exclusively for VIO and the other is the narrow FOV
camera, that is used for matching against Street View imagery. It is important
to note that they point in different directions and do not share views. For
this reason, the resulting features for VIO and 3D localization are not
directly correlated. Note also that our two step optimization can in principle
be done in one step. Our choice to do it in two steps resulted from a
practical perspective: the first is used to compute a good initial solution for
the second optimization.  For the scope of this paper, we are not interested in
using the panoramas to build an accurate large model of the environment: we aim
at localizing without building new large scale maps where Street View exists.

% Since we do not connect multiple panoramas to make the system less
% reliant on finding multiple panorama matches, performing a single joint
% optimization is not helpful.
% this make possible to the system online.

\section{Conclusion}

In this paper, we present a novel approach to metric localization by
matching Google's Street View imagery to a moving monocular camera.
Our method is able to metrically localize without requiring a robot to
pre-visit locations to build a map where Street View exists.

We model the problem of localizing a robot with Street View imagery as
a non-linear least squares estimation in two phases. The first
estimates the 3D position of tracked feature points from short
monocular camera streams, while the second computes the rigid body
transformation between the points and the panoramic image.  The sensor
requirements of our technique are a monocular image stream and
odometry estimates. This makes the algorithm easy to deploy and
affordable to use. In our experiments, we evaluated the metric
accuracy of our technique by using fiducial markers in a wide outdoor
area.  The results demonstrate high accuracy in different
environments. Additionally, to show the flexibility and the potential
application of this work to personal localization, we also ran
experiments using images acquired with Google Tango smartphone in two
different urban scenarios. We believe that this technique paves the
way towards a new cheap and widely useable outdoor localization
approach.

\small
\bibliography{robotics}
\bibliographystyle{abbrvnat}
\end{document}